\documentclass[runningheads]{llncs}
\usepackage{graphicx}
\usepackage[symbol]{footmisc}
\usepackage{amsmath,amssymb} 
\usepackage{color}
\usepackage[utf8]{inputenc} 
\usepackage{hyperref}       
\usepackage{url}            
\usepackage{booktabs}       
\usepackage{amsfonts}       
\usepackage{nicefrac}       
\usepackage{microtype}      
\usepackage{algorithm}
\usepackage{algorithmic}

\begin{document}
\title{Efficient Automatic Meta Optimization Search for Few-Shot Learning}
%

\newcommand*\samethanks[1][\value{footnote}]{\footnotemark[#1]}

\author{Xinyue Zheng \thanks{These authors contributed equally to this work} \and
Peng Wang \samethanks \thanks{Corresponding author: Peng Wang} \and
Qigang Wang \and 
Zhongchao shi \and 
Feiyu Xu}

\authorrunning{Xinyue Zheng et al.}
%
\institute{AI Lab, Lenovo Research, Beijing 100089, China \\
\email{\{zhengxy7, wangpeng31, wangqg1, shizc2, fxu\}@lenovo.com}}

\maketitle              
\begin{abstract}

Previous works on meta-learning either relied on elaborately hand-designed network structures or adopted specialized learning rules to a particular domain. We propose a universal framework to optimize the meta-learning process automatically by adopting neural architecture search technique (NAS). NAS automatically generates and evaluates meta-learner's architecture for few-shot learning problems, while the meta-learner uses meta-learning algorithm to optimize its parameters based on the distribution of learning tasks. Parameter sharing and experience replay are adopted to accelerate the architectures searching process, so it takes only $1$-$2$ GPU days to find good architectures. Extensive experiments on Mini-ImageNet and Omniglot show that our algorithm excels in few-shot learning tasks. The best architecture found on Mini-ImageNet achieves competitive results when transferred to Omniglot, which shows the high transferability of architectures among different computer vision problems.

\end{abstract}
\section{Introduction}

Many meta-learning methods \cite{LarochelleOPTIMIZATION,Memory-Augmented,Matching} have achieved success in “$K$-shot, $N$-way” scenario. In this scenario, each task is a $N$-classification problem, and the learner only sees $K$ training instances from each class. After training with these training instances, the learner is able to classify new images in the test set. Finn et al. \cite{MAML} proposed a model-agnostic meta-learning approach (MAML). The key breakthrough is its initialization technology, which allows the learner to repeatedly train on each sampled task and set parameters at the optimal start point. Compared to MAML, which needs to calculate second derivatives in back-propagation, Reptile \cite{reptile} only uses first-order derivatives with higher efficiency and less computational resource. However, the Reptile algorithm only optimizes meta-learner from the parameters level, and the learner's model is a simple and powerful network structure that are artificially designed. Designing architectures is a time-consuming process which often requires rich expert knowledge and many experimental comparisons. Therefore, we propose a novel joint optimization scheme which combines model-agnostic meta-learning algorithm and automatic architecture design to improve the few-shot learning.

As is shown in Figure 1., our scheme employs the neural architecture search technique to automate architecture design process. The contributions of each component are listed as follows: The controller is trained with policy to sample the meta-learner’s architecture from component library; meta-learner uses Reptile to seek high adaptive initial parameters for different tasks; experience replay and parameter sharing speed up meta-learner search by learning from historical knowledge.

To be specific, our model search method is based on ENAS\cite{pham2018efficient} which improves the efficiency of NAS by allowing parameter sharing among generated model. Figure 1. shows a recurrent network – the controller that outputs variable-length string to define a child model, including configurable model depth, stochastic skip connection, and different combination of convolution cells. The parameters of child model are trained by Reptile \cite{reptile}. After a period of Reptile training, child model returns the accuracy as reward to evaluate and adjust the controller's architecture-generation policy. To speed up model searching procedure, we apply experience replay to reduce the number of controller interactions with the environment and encourage the controller to fully study its accumulated experience in the changing environment. Ultimately, the controller can optimize its policy and yields the best model architecture. We retrain this model from scratch using Reptile, and it can generalize across tasks with only a small number of gradient steps using few samples on each task.

\begin{figure*}[!tp] 
\centering
\includegraphics[width=4.7in]{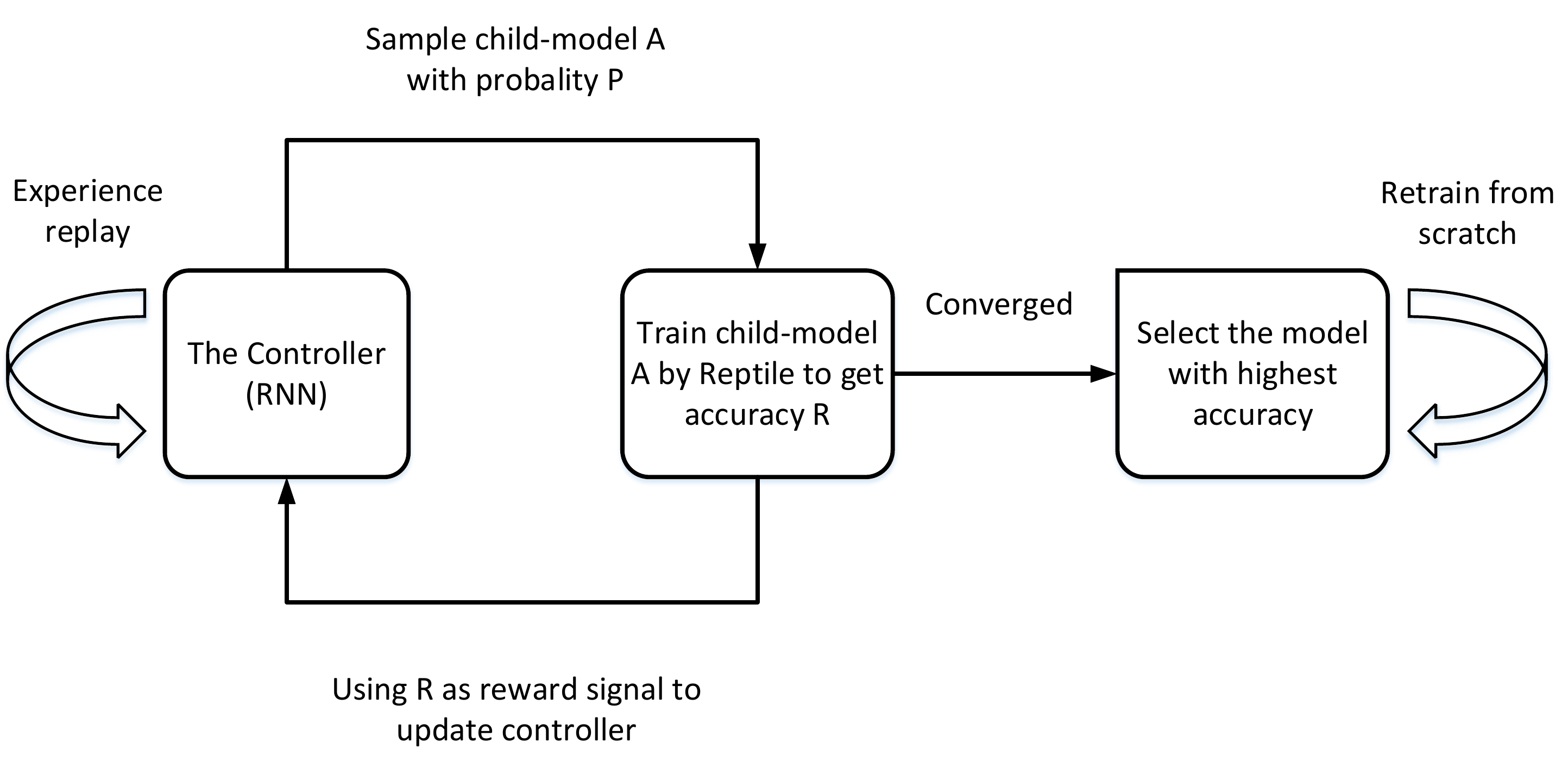}
\caption{An overview of efficient automatic meta-learning method}
\end{figure*}

We make the following contributions:
\begin{itemize}
\item We are the first to propose an automatic meta-optimization system by applying neural architecture search technique to meta-learning method.

\item A series of experiments show that the joint automatic optimization method can ensure that the training model has rich expression ability and high cross-task generalization ability. On Mini-ImageNet benchmark, it achieves excellent performance with $74.20 \%$ accuracy.

\item We achieve remarkable meta-learner search efficiency ($5$-shot with $48$ GPU hours; $1$-shot with $32$ GPU hours.). It credits to the incorporation of parameter sharing and experience replay techniques in search process.

\item The algorithm shows high transferability among different computer vision problems. The best architecture found on Mini-ImageNet achieves competitive results on Omniglot tasks, and the searched models in $5$-shot, $5$-way classification are transferable to $1$-shot, $5$-way scenario. 

\end{itemize}

\section{Related Work} 

\subsection{Meta Learning}

Meta-learning allows learners to train through a variety of similar tasks, and expects to generalize to previously unseen tasks quickly. There are several ways to realize meta-learning. Memory based methods \cite{meta,Memory-Augmented} adjust bias by weights update and generate outputs by learning from memories. Santoro et al. \cite{Memory-Augmented} make use of external memory introduced by Neural Turing Machine \cite{Turing} to realize short term memory and build connections between labels and input images, so that latter inputs are able to compare with related images in memories to achieve better predictions. Gradient based methods \cite{Learning,LarochelleOPTIMIZATION} train a LSTM optimizer to learn parameter optimization rules of the learner network. While \cite{Learning} targets at large-scale classification, \cite{LarochelleOPTIMIZATION} is interested in few-shot learning and learns both optimization rules and weight initialization. Recent work such as relation network \cite{Relation} and matching network \cite{Matching} employ idea from metric learning. Instead of using artificially designed metrics, it completely utilizes neural networks to learn deep distance metric. Simple Neural Attentive Learner (SNAIL) \cite{SNAIL} uses temporal convolution to collect past experience and soft attention to pinpoint specific pieces of details. Object-level representation learning \cite{Object} decomposes images into objects, and applies object-level relation learned from source dataset to the target dataset.

Although the existing approaches have achieved impressive results, they either introduce extra parameters which need more storage spaces or bring constraints on the model architecture. MAML \cite{MAML} is well accepted for its simplicity and model-agnostic. This method learns highly adaptive parameters to initialize the neural network so that only a small number of gradients updates are required for fast learning on a new task. Recently, OpenAI proposes a similar method Reptile \cite{reptile} which does not require differentiability during the optimization process compared to MAML.

\subsection{Neural Architecture Search}

Human-designed networks usually only perform specific tasks. An automated method to generate appropriate architecture with adaptive model parameters and hyperparameters for any given tasks is desired. Many hyperparameter optimization methods have been studied \cite{snoek2012practical,li2016hyperband,loshchilov2016cma,hazan2017hyperparameter}. These optimization algorithms are able to select and fine tune the model hyperparameters automatically which surpass human expert-level optimizations. However, they are not flexible and often limited in generating fixed-length configuration for networks.

Recent years evolutionary algorithms and reinforcement learning algorithms have been adopted for neural architecture search and achieved promising performance. Neuro-evolution methods \cite{real2017large,real2018regularized} use mutation operations to explore large search spaces, which have expensive evaluation cost and need heuristic algorithms. The reinforcement learning approach has higher feasibility and achieves better results. Zoph et al. \cite{zoph2016neural,zoph2017learning} use recurrent network to generate expected "child model", and utilize the accuracy of the child model on the validation set as reward signals to train this RNN. Efficient Neural Architecture Search (ENAS) \cite{pham2018efficient} speeds up the training process by allowing parameter sharing among child models. Another efficient method Differentiable Architecture Search (DARTS) \cite{DARTS} constructs continuous search space and optimizes architecture in a differentiable manner.

\section{Method}

In meta-learning, learners make progress at task level rather than data point level. For example, MAML \cite{MAML} takes in a distribution of tasks, where each task is an independent learning problem. In order to lower the loss $L_{\tau}$ on task $\tau$, we need to compute the following formula:

\begin{equation}
\min_{\theta, \mathcal{A}}\sum_{\tau}{L(D_{\tau}^{'},\theta_{\tau}^{'})}=\sum_{\tau}{L(D_{\tau}^{'}, T(D_{\tau},\theta))}
\end{equation}

where $D_{\tau}$ and $D_{\tau}^{'}$ represent the training set and test set on task $\tau$ respectively, the $T(D_{\tau},\theta)$ is the training procedure acting on $D_{\tau}$, and the $L_{\tau}$ is computed on updated parameters $\theta'$ with test samples $D_{\tau}^{'}$. $\mathcal{A}$ represents the model architecture of the meta-learner, and the $\theta$ are the model parameters under this architecture. In classic meta-learning setting, $\mathcal{A}$ is fixed and we only optimize $\theta$. In our proposed scheme, $\mathcal{A}$ and $\theta$ will be joint optimized with the alternatively training manner.       

There are two stages in each meta-optimization search step. First the controller with policy $\phi$ is trained to sample a architecture $\mathcal{A} = f(\phi, R)$, where $R$ is the reward output by the meta-learner, it is a random value at the first time. Second, using this architecture, the meta-learner is trained with Reptile algorithm. Reptile \cite{reptile} is the first order of MAML. Using the same principle, it seeks an initialization condition for model parameters which can be fine-tuned easily, so that the trained learner is able to achieve high performance on previously unseen tasks. The score on validation set will be input as reward to the controller based on reinforcement learning. These two stages are alternatively trained until some good architecture candidates are generated. After all search steps finished, we would finally retrain these candidates to obtain the architecture with highest score on the meta-test dataset. 

Since the discrete domain search of $\mathcal{A}$ can be transformed into the continuous domain optimization of controller network in our method, the formula (1) can be rewrite as a differential form which can be optimized with end-to-end training:

\begin{equation}
\min_{\theta, \phi}\sum_{\tau}{L(D_{\tau}^{'},\theta_{\tau}^{'})}=\sum_{\tau}{L(D_{\tau}^{'}, T(D_{\tau},\theta))}
\end{equation}

\subsection{Generating Transferable Architecture by Controller}
We use LSTM as the controller to generate a variable-length string which specifies the model architecture. The controller receives a randomly initialized variable as input at the very beginning, then the input of time step $t$ is the embedding of the decision sampled from time step $t-1$.

As shown in Figure 2, the controller aims to generate a four-layer child model architecture. It needs to make two sets of decisions according to the current generation policy: 1) what operations to be applied and 2) which previous layers to be concatenated. There are several available operations: ordinary convolutions, depthwise-separable convolutions, average pooling and max pooling. After selecting the operation, the controller will decide to select the previous skip connection layers. Take layer $k$ as an example, $k-1$ indices of previous layers are sampled, which conduce to $2^{(k-1)}$ possible choice. Corresponding to Figure 2, at layer $k=3$, the controller selects the indices ${1,2}$, which means the output of layer $1$ and $2$ are concatenated along depth dimension and sent to layer $4$.

\begin{figure*}[!tp] 
\centering
\includegraphics[width=4.9in]{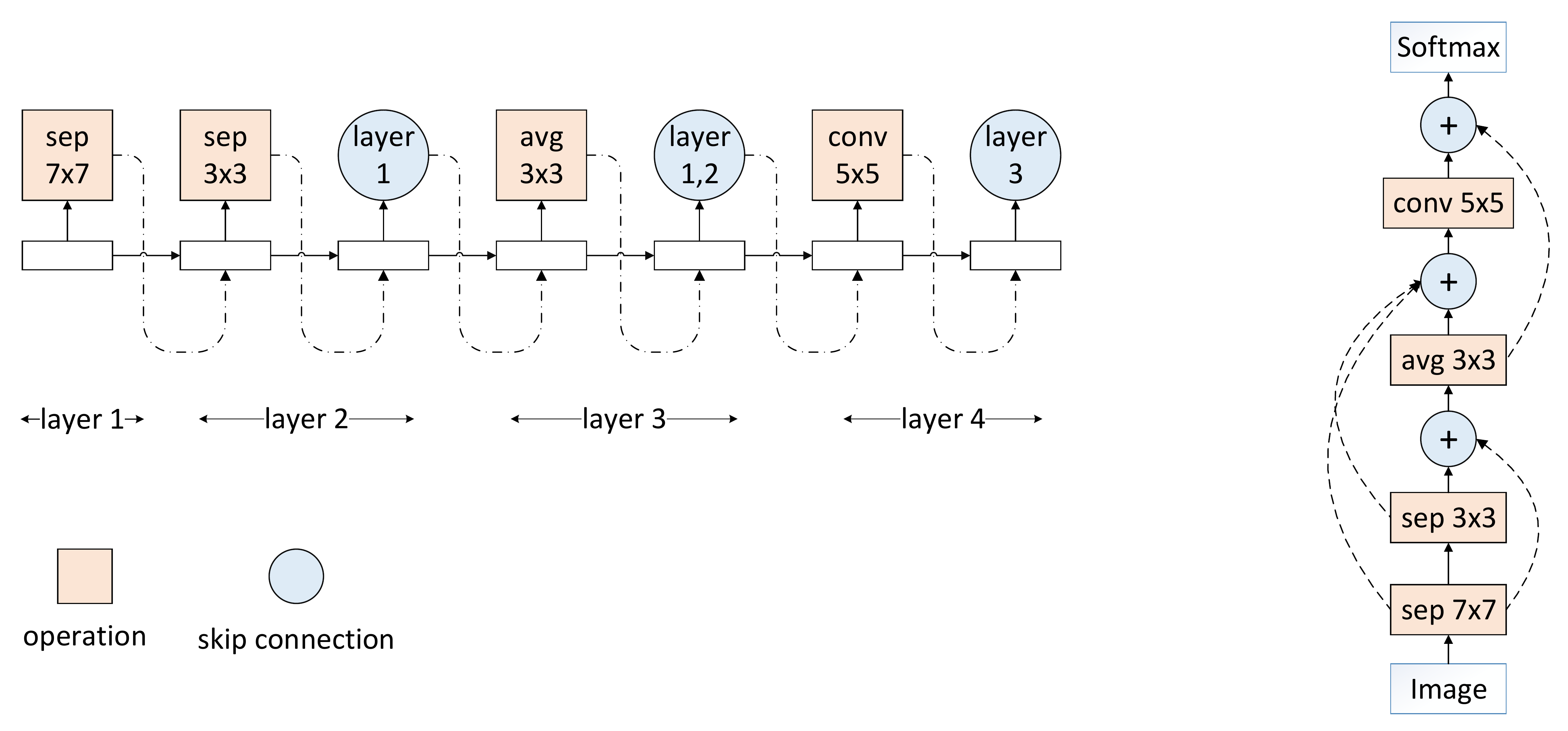}
\caption{\emph{Left}: The prediction string made by controller. \emph{Right}: Connect prediction string to build a complete network.}
\end{figure*}

\subsection{Training Controller with Reinforcement Learning}

\subsubsection{Policy gradient}
When training the controller, we freeze parameters of the child model and only update controller's policy $\phi$  (referred to as Algorithm $1$).  Actions $a_{1:T}$ are the decisions made by the controller's policy in time series: selecting an operand and layers for skip connection. We utilize policy gradient \cite{Sutton1999Policy} to train the controller to maximize the expected reward $E_{P(a_{1:T}; \phi)} [R]$. The traditional policy gradient formula is: 
\begin{equation}
\nabla_{\phi}U\left(\phi\right)=\frac{1}{m}\sum_{j=1}^m\sum_{t=1}^T{\nabla_{\phi}\log P\left(a_t|a_{(t-1):1} ;\phi\right)R_j}
\end{equation}

where $m$ stands for the $m$ architectures sampled by the current policy. $T$ denotes the number of predictions made by the controller. $R$ controls the parameter update direction and step size. Equation ($3$) targets at increasing the generation probability of high reward models and reducing the opposite. We employ the empirical average reward of these $m$ architectures to approximate the policy gradient. 

The above method is unbiased but with high variance. As is proposed by \cite{zoph2016neural}, we introduce a baseline $bl$ in reward to reduce the variance. $bl$ is defined as the exponential moving average of previous architecture accuracy. By subtracting the baseline, we can understand the improvement of a model compared with an average one.
\begin{equation}
A_j = R_j - bl_j
\end{equation}
\begin{equation}
\nabla_{\phi}U\left(\phi\right)=\frac{1}{m}\sum_{j=1}^m\sum_{t=1}^T{\nabla_{\phi}\log P\left(a_t|a_{(t-1):1};\phi\right)A_j} 
\end{equation}
The advantage function $(4)$ helps to update policy parameters $(5)$ with a more specific direction.

\begin{algorithm}[!h]  
    \caption{Automatic architecture search with experience replay.}  
    \begin{algorithmic}[1]  
        \STATE Randomly initialize policy $\phi$, input state $S_0$ 
        \FOR {$j = 1$ to $J$}
        \STATE Observe $S_j$, and generate action stream $A_j=a_{1:T}$ to form a child model
        \STATE Train Child model with \textbf{Algorithm 2} on meta-training to get reward $R_j$ 
        \STATE Perform PG to $\phi$ with this experience ($S_j, A_j, R_j$)
        \STATE Store ($S_j, A_j, R_j$) if $R_j \ge bl$, or with probability $\frac{R_j}{bl}$ into replay buffer $M$    
        \IF {$j \bmod k$ == $0$}
        \FOR {$r = 1$ to $R$}
        \STATE Uniformly sample transition $B$ from $M$  
        \STATE Update policy parameters $\phi$ with $B$
        \ENDFOR 
        \ENDIF
        \ENDFOR
             
    \end{algorithmic}  
\end{algorithm}

\subsubsection{Experience Replay}

Policy gradient is based on stochastic gradient algorithms. The controller is updated using only one sample architecture generated by the current policy and discards it after a single update. Therefore, the controller tends to forget its past experience which leads to oscillation. We solve this issue by adopting experience replay skills \cite{PrioritizedEP}. To perform experience replay, we store transition $(S_j, A_j, R_j)$  in a replay buffer, where $A_j$ stands for the $j$-th architecture string $a_{1:T}$ selected by the controller, $S_j$ refers to the input state of the controller, and $R$ corresponds to the accuracy computed on the validation set. Since not all experiences are expected to be learned more than once, experience will be stored in the buffer with probability:

\begin{equation}
P_j = \left\{
             \begin{array}{lc}
             \frac{R_j}{bl},  & R_j \le bl_j     \\
             1,  & R_j > bl_j     
             \end{array}  
        \right.
\end {equation}
The criterion to measure the importance of transition is its reward $R_j$, which suggests how good this architecture is compared with current baseline $bl_j$.

\subsection{Training Child Model with Reptile}

When training the child model (referred to as Algorithm $2$), we freeze the controller's policy parameters $\phi$. Child models are required to learn from limited number of images, so we build our work on a scalable meta-learning algorithm Reptile. Assume that $p(T)$ are the distribution probability  of tasks, we sample a batch of tasks $T$ from $p(T)$. The standard cross-entropy loss $L_{\tau_i}$ denotes the task-specific loss function. In order to make the model parameters $\theta$ sensitive, we calculate each task $\tau_i$ with $k$ gradient steps on loss $L_{\tau_i}$ and get the final parameter vector $W_i$. Meta-optimization across tasks is performed via Adam algorithm, where $\sum\nolimits_{i=1}^N(W_i - \phi)$ is treated as gradient. Besides, training the child model on the validation set generates accuracy $R$, which will be returned as the reward to scale gradients of the controller.

\begin{algorithm}[!h]  
    \caption{Reptile at training time.}  
    \begin{algorithmic}[1]  
        \STATE Randomly initialize $\theta$ 
        \REPEAT
        \STATE Sample batch of $N$ tasks $T_b \sim p(T)$
        \FOR {each $\tau_i$ in $T_b$}  
        \STATE Compute adapted parameters with $k$ gradients step: $W_i = SGD(L_{\tau_i}, \theta, k)$ 
        \ENDFOR 
        \STATE Update $\theta \leftarrow \theta + \alpha\frac{1}{k}\sum\nolimits_{i=1}^N(W_i - \theta)$
        \UNTIL {Convergence}
        \RETURN {Accuracy on validation set as reward $R_j$}
    \end{algorithmic}  
\end{algorithm}

\section{Experiment}
In this section, we evaluate automatic meta-learning method on two important benchmarks: Mini-ImageNet and Omniglot, and compare our results against strong baselines. All of our experiments consider solving $K$-shot, $N$-way learning problem. For each task $\tau_i$ of $K$-shot, $N$-way classification, the learner trains on $N$ related classes each with $K$ examples, we firstly sample $N$ classes from meta-dataset and then select $K+1$ examples for each class. Then, we split these examples into training and test sets, where training set $D_{train}$ contains $K$ examples for each class and test set $D_{test}$ contains the remaining sample. Take $5$-shot, $5$-way classification as example, we use $25$ examples --- $5$(images) x $5$ (classes) to train the learner and use additional examples to test the model.

\subsection{Few-shot Learning Datasets}
Mini-ImageNet is created by randomly sampling $100$ classes from ImageNet and selecting 600 examples for each class. Training set has $38,400$ images with $64$ classes, test set consists of $12,000$ images with $20$ classes, and validation set contains $9600$ images with $16$ classes.

Omniglot consists of $1623$ characters from $50$ different alphabets. We randomly select 1200 characters for training and use the remaining character classes for testing. As is proposed by Santoro et al. \cite{Memory-Augmented}, we augment the dataset with rotations by multiples of $90$ degrees. Omniglot was proposed by lake and used in the 2015 Science paper \cite{Lake2015Human}.

\subsection{Implementation Details}

The controller is a one-layer LSTM with 100 hidden units, whose goal is to search $8$-layers child models. Operations can be selected from: $3$ x $3$, $5$ x $5$, and $7$ x $7$ convolutions, $3$ x $3$ , $5$ x $5$, and $7$ x $7$ depthwise-separable convolutions and $3$ x $3$ average pooling and max pooling. We perform experience replay on the controller every $60$ steps and $5$ transitions each time. A global average pooling is added before the fully connected layer and dropout layers with $0.25$ drop rates are added after each layer. These tricks reduce the number of parameters and avoid overfitting during training. We use $1$ GPU for $1$-$2$ days to search for top-$3$ architectures for meta-learner, and each architecture takes $6$ GPU hours to retrain from scratch. Table $1$ presents parameter settings in the final retrain process.

\begin{table}[!h]
\renewcommand\arraystretch{1.1} 
\centering  
\caption{Parameters for final-retrain on Mini-ImageNet}
\begin{tabular}{lccc}  
\hline
Parameters  &$5$-shot $5$-way & $1$-shot $5$-way \\ 
\hline
Adam learning rate &$0.005$ &$0.003$ \\
Outer iterations &$7K$ &$7K$ \\
Outer step size &$1$ &$1$ \\
Meta batch size &$5$ &$5$ \\
\hline
Inner iterations &$8$ &$8$ \\
Inner batch size &$10$ &$10$ \\
Train shots &$15$ &$15$ \\
\hline
Eval. inner iterations &$88$ &$50$ \\
Eval. inner batch size &$10$ &$5$ \\
\hline
\end{tabular} 
\end{table}

\subsection{Evaluation}

\begin{table*}[!h]
\renewcommand\arraystretch{1.2} 
\renewcommand\tabcolsep{10.0pt}
\centering  
\caption{Results on Mini-ImageNet}
\begin{tabular}{lccr}  
\hline
Algorithm & Transduction & $5$-shot $5$-way & $1$-shot $5$-way\\ 
\hline

MAML\cite{MAML}  & Y & $63.11 \pm 0.92 \%$ &$48.70 \pm 1.84 \%$ \\
Reptile\cite{reptile} &  N & $62.74 \pm 0.37 \%$  & $45.79 \pm 0.44 \%$ \\
Reptile & Y & $66.00 \pm 0.62 \%$ & $48.21 \pm 0.69 \%$ \\
Matching Nets\cite{Matching}& N & $55.31 \pm 0.73 \%$ &$43.56 \pm 0.84 \%$ \\
Relation Nets\cite{Relation} & N & $65.32 \pm 0.70 \%$ &$50.44 \pm 0.82 \%$ \\
SNAIL\cite{SNAIL} & N & $68.88 \pm 0.92 \%$ & $\textbf{55.71} \pm \textbf{0.99\%}$ \\

\hline
Ours  & N & $67.10 \pm 0.90 \%$ & $48.00 \pm 0.82\%$\\
\textbf{Ours}  & Y &  $\textbf{74.20} \pm \textbf{0.32\%}$  & $\textbf{52.43} \pm \textbf{1.08\%}$ \\
Ours (Transfer)  & N & $\backslash$  &$47.04 \pm 0.52\%$ \\
Ours (Transfer) & Y & $\backslash$ &$51.62 \pm 0.43\%$ \\
\hline
\end{tabular} 
\end{table*}

As shown in Table $2$, Our method achieves competitive results on Mini-ImageNet. In transductive mode, the trained model classifies all the samples in test set at once, so the information is allowed to leak between test samples through batch normalization \cite{reptile}. As expected, the transductive experiments achieve $74.2\%$ high accuracy on $5$-shot $5$-way task.

Automatic searching process learns directly from the task distribution of dataset, We show it enables some degree of transferability. In experiments, we transfer the model constructed from $5$-shot $5$-way configuration into $1$-shot $5$-way for final-retrain. It achieves $51.62\%$ accuracy which still beyond the original Reptile performance on Mini-ImageNet.

Although some methods \cite{Relation,SNAIL} have achieved competitive performance as ours, our method sacrifices some accuracy in exchange for time and space efficiency. For example, we only select the top3 searched architectures for retrain. If we select top10, or top100 architectures for retrain, the accuracy may be improved. In addition, the network could automatically select the number of layers, the number of feature maps, etc., which is easier than the manual setting to find the most powerful architecture, but this procedure is very time-consuming. What's more, we use the experience replay to encourage learner learning from past experience but reduce the number of times to explore new architectures. Although it greatly improves the efficiency, it reduces the possibility of exploring the best architecture to some extent. Therefore, if we only focus on accuracy, there is a great room for improvement, but we think efficient algorithm with competitive results are more valuable. Compared to the original NAS technology, who takes 32,400-43,200 GPU hours, our algorithm can search for good architectures within 48 GPU hours.

\begin{table*}[!h]
\renewcommand\arraystretch{1.2} 
\renewcommand\tabcolsep{7.0pt}
\centering  
\caption{Results on Omniglot}
\begin{tabular}{lccc}  
\hline
Algorithm & Transduction & $5$-shot $20$-way & $1$-shot $20$-way\\ \hline
Matching Nets\cite{Matching} & N & $98.7 \%$ & $93.5 \%$ \\
$1^{st}$order MAML\cite{MAML} & Y & $97.0 \pm 0.1 \%$ & $89.4 \pm 0.5 \%$\\
MAML & Y & $98.9 \pm 0.2 \%$ &$95.8 \pm 0.3 \%$ \\
Reptile\cite{reptile} & N &  $96.65 \pm 0.33 \%$  & $88.14 \pm 0.15 \%$ \\
Reptile & Y &  $97.12 \pm 0.32 \%$ & $89.43 \pm 0.14 \%$ \\
\hline
Ours & N & $98.97 \pm 0.12 \%$ & $95.50 \pm 0.35\%$\\
Ours (Transfer)  & N & $97.95 \pm 0.23\%$  &$93.80 \pm 0.18\%$ \\
\hline
\end{tabular} 
\end{table*}

In Omniglot, we try the distance transfer experiment to test the generalization performance of the searched architecture. 
Here, we merely transfer the model architecture from Mini-ImageNet, but all the weights will be re-trained from scratch. From Table 3. we find that transferred architecture generalize well to Omniglot problems, even exceeds the accuracy of original Reptile method with transductive setting. So the automatic meta-learning method has been proven not only to achieve within-task generalization, but also cross-task generalization.

Figure 3. shows the experience replay contributes to the understanding of new tasks, and helps the learner to discover better architectures with less computational cost. Note that the y-axis is the moving average of previous architecture accuracy, so it only reflects the average value and many architectures can achieve much higher accuracy. Figure 4. shows the good architectures we discovered.

\begin{figure}[h] 
\centering
\includegraphics[width=3.5in]{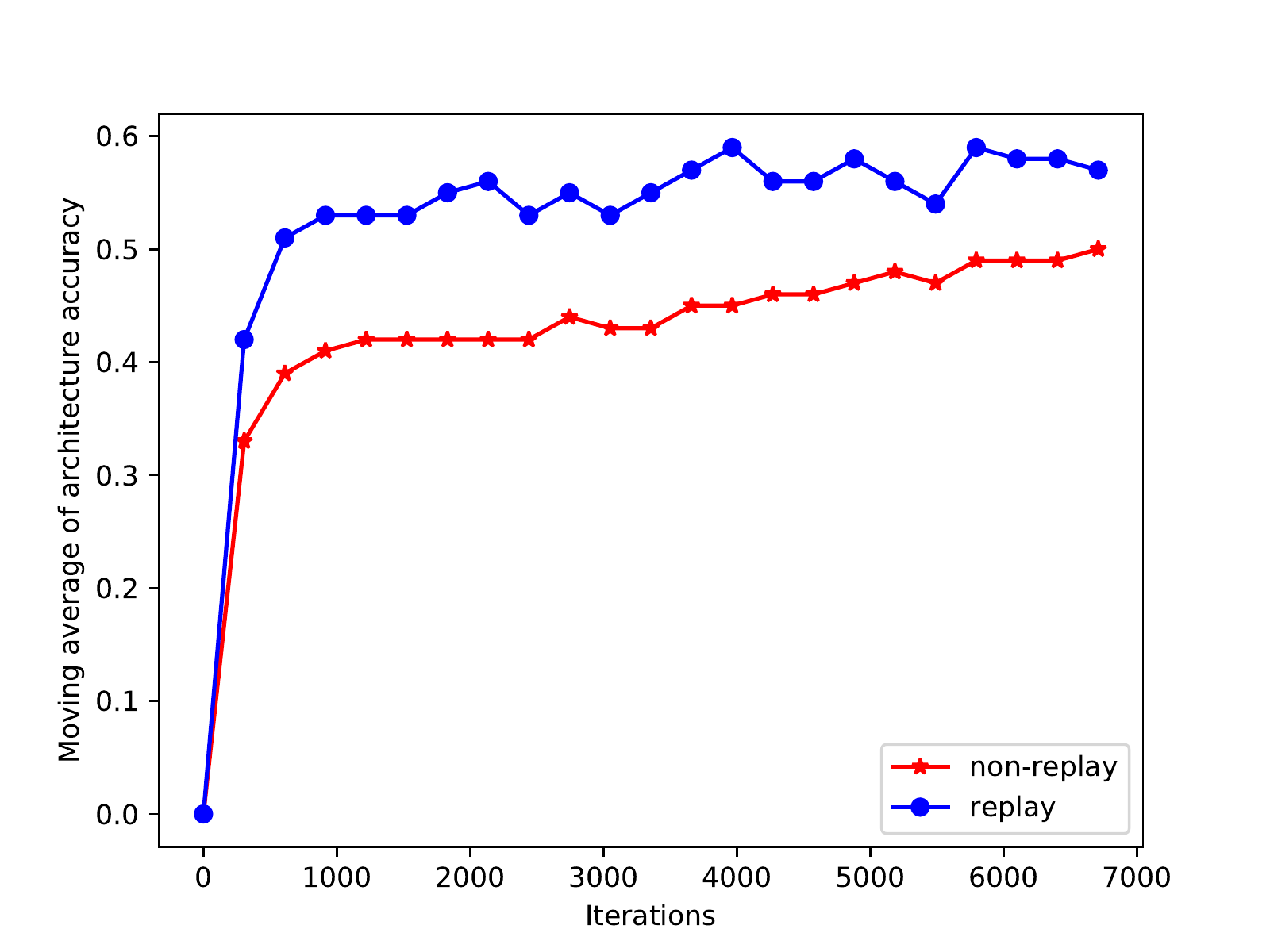}
\caption{Training curves for the architecture search procedure: exponential moving average architecture accuracy over $7$K iterations.}
\end{figure}

\begin{figure*}[!h] 
\centering
\includegraphics[width=4.8in]{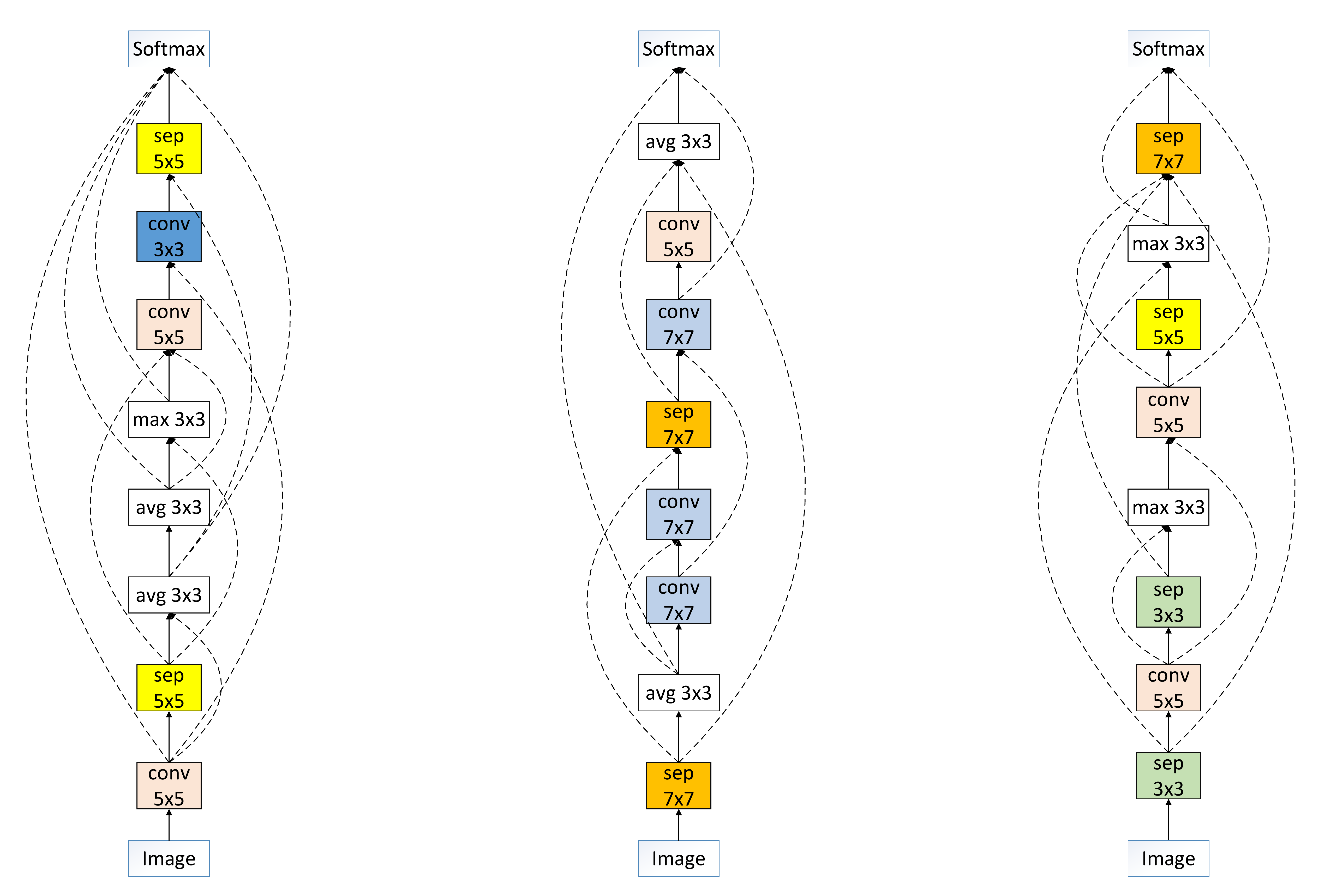}
\caption{High accuracy architectures searched by $5$-shot, $5$-way classification on Mini-ImageNet, and can be transferred to other classification scenes.}
\end{figure*}

\section{Conclusion}

In this paper, we introduce an efficient automatic meta optimization search for few-shot learning problems. Rich experiments show that the proposed algorithm achieves competitive performance in few-shot learning tasks. Our work has a few key insights. Firstly, the proposed framework is universal because the architecture search will discover scalable architectures for meta-learner, which can be easily nested on any model-agnostic meta-learning algorithm. Secondly, parameter sharing and experience replay techniques greatly save the computational cost and improve the efficiency of our approach. Lastly, We show the within-task generalization and cross-task generalization of the learner's architecture, this transferability is a desirable characteristic and deserves further study.

\bibliographystyle{splncs04}
\bibliography{egbib}

\end{document}